\documentclass[lettersize,journal]{IEEEtran}
\usepackage{amsmath,amsfonts}
\usepackage{algorithmic}
\usepackage{algorithm}
\usepackage{array}
\usepackage[caption=false,font=normalsize,labelfont=sf,textfont=sf]{subfig}
\usepackage{textcomp}
\usepackage{stfloats}
\usepackage{url}
\usepackage{verbatim}
\usepackage{graphicx}
\usepackage{cite}
\usepackage{multirow}
\usepackage{comment}
\usepackage{color}
\usepackage{float}
\usepackage{fancyhdr}

\begin{document}

\title{Weight-dependent Gates for Network Pruning}


\author{Yun Li, Zechun Liu, Weiqun Wu, Haotian Yao, Xiangyu Zhang, Chi Zhang, and Baoqun Yin
\thanks{
Yun Li and Baoqun Yin are with University of Science and Technology of China, Hefei, China (e-mail: yli001@mail.ustc.edu.cn, bqyin@ustc.edu.cn). (Corresponding author: Baoqun Yin.)

Zechun Liu is with Hong Kong University of Science and Technology, Hong Kong, China (e-mail: zliubq@connect.ust.hk).

Weiqun Wu is with Chongqing University, Chongqing, China (e-mail: wuwq@cqu.edu.cn)

Haotian Yao, Xiangyu Zhang, and Chi Zhang are with Megvii Inc., Beijing, China (email: yaohaotian@megvii.com, zhangxiangyu@megvii.com, zhangchi@megvii.com).

}
}

\markboth{Journal of \LaTeX\ Class Files,~Vol.~14, No.~8, August~2021}%
{Shell \MakeLowercase{\textit{et al.}}: A Sample Article Using IEEEtran.cls for IEEE Journals}


\maketitle





\begin{abstract}
In this paper, a simple yet effective network pruning framework is proposed to simultaneously address the problems of pruning indicator, pruning ratio, and efficiency constraint. This paper argues that the pruning decision should depend on the convolutional weights, and thus proposes novel weight-dependent gates (W-Gates) to learn the information from filter weights and obtain binary gates to prune or keep the filters automatically. To prune the network under efficiency constraints, a switchable Efficiency Module is constructed to predict the hardware latency or FLOPs of candidate pruned networks. Combined with the proposed Efficiency Module, W-Gates can perform filter pruning in an efficiency-aware manner and achieve a compact network with a better accuracy-efficiency trade-off. We have demonstrated the effectiveness of the proposed method on ResNet34, ResNet50, and MobileNet V2, respectively achieving up to 1.33/1.28/1.1 higher Top-1 accuracy with lower hardware latency on ImageNet. Compared with state-of-the-art methods, W-Gates also achieves superior performance. 
\end{abstract}

\begin{IEEEkeywords}
Weight-dependent gates, switchable Efficiency Module, Accuracy-efficiency trade-off, Network pruning
\end{IEEEkeywords}

\section{Introduction}

\IEEEPARstart{I}{n} recent years, convolutional neural networks (CNNs) have achieved state-of-the-art performance in many tasks, including but not limited to image classification \cite{he2016deep, wang2016cost}, semantic segmentation \cite{zhuang2020video}, and object detection \cite{wang2021dynamic, girshick2014rich}, etc. Despite their great success, billions of float-point-operations (FLOPs) cost and long inference latency are still prohibitive for CNNs to deploy on many resource-constraint hardware. As a result, a significant amount of effort has been invested in CNNs compression and acceleration, in which filter pruning \cite{guo2020model,li2016pruning} is seen as an intuitive and effective network compression method.

However, filter pruning is non-trivial and faces three major challenges. 1) Pruning indicator: CNNs are usually seen as a black box, and individual filters may play different roles within and across different layers in the network. Thus, it is difficult to manually design indicators that can fully quantify the importance of their internal convolutional filters and feature maps. 2) Pruning ratio: How many filters should be pruned in each layer? The redundancy varies from different layers, making it a challenging problem to set appropriate pruning ratios for different layers. 3) Efficiency constraint: Most previous works only adopt hardware-agnostic metrics such as parameters or FLOPs to evaluate the efficiency of a CNN. But the inconsistency between hardware-agnostic metrics and actual efficiency \cite{Sandler_2018_CVPR} lead to an increasing industrial demand on directly optimizing the hardware latency. 

Previous works have tried to address these issues from different perspectives. Conventional filter pruning works mainly rely on manual-designed indicators \cite{li2016pruning, he2017channel, li2019exploiting} or data-driven indicators \cite{huang2018data, liu2017learning}. However, manual-designed indicators usually involve human participation, and data-driven pruning indicators may be affected by the feature maps. Besides, in previous works, the pruning ratio of each layer or a global pruning threshold is usually human-specified, making the results prone to be trapped in sub-optimal solutions\cite{liu2019metapruning}.

\begin{figure*}[t]
	\begin{center}
		\includegraphics[width=1.0\linewidth]{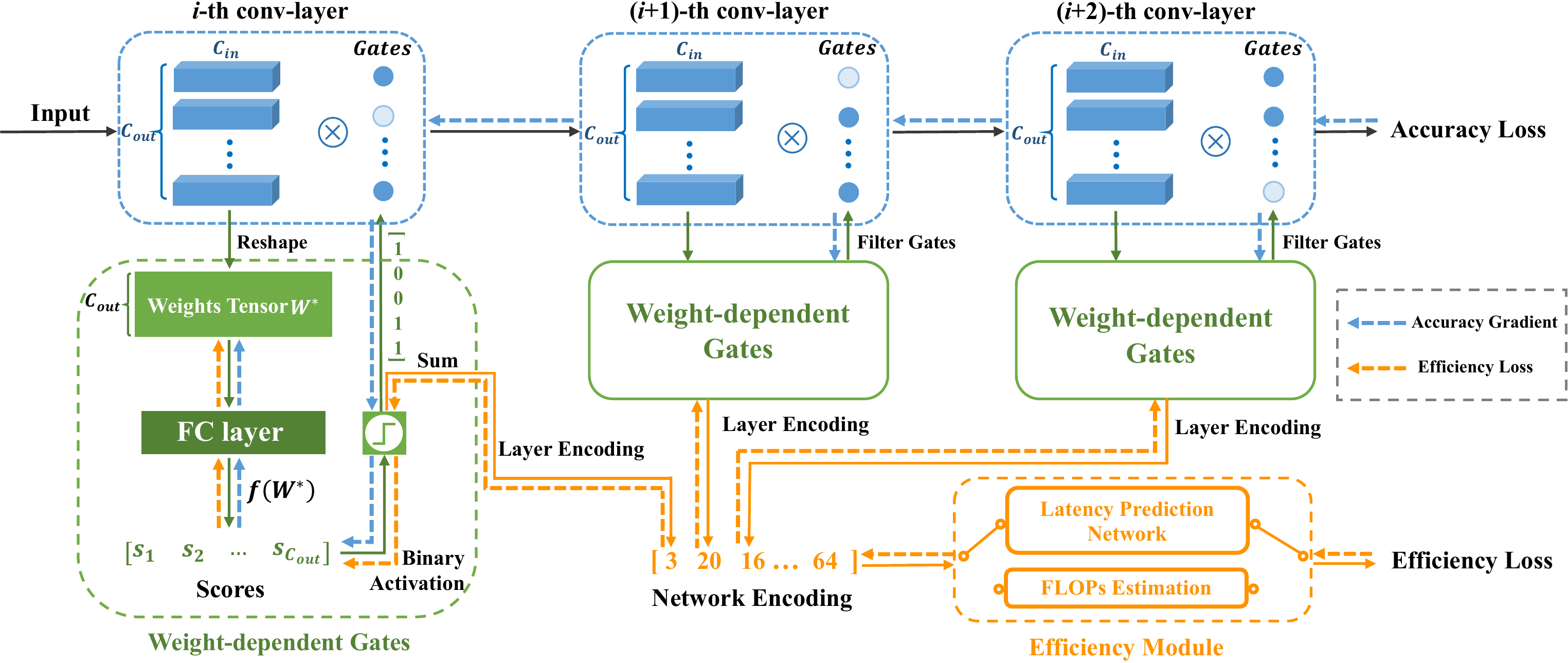}
	\end{center}
	\caption{Pruning framework overview. There are two main parts in our framework, Weight-dependent Gates (W-Gates) and a switchable Efficiency Module. 
	The W-Gates learns the information from the filter weights of each layer and generates binary filter gates to open or close corresponding filters automatically (0: close, 1: open). The filter gates of each layer are then summed to obtain a layer encoding and all the layer encodings constitute a network encoding. Next, the network encoding of the candidate pruned network is fed into Efficiency Module to get the predicted latency or FLOPs of the candidate pruned network. During the pruning process, accuracy loss and efficiency loss compete against each other to obtain a compact model with a better accuracy-efficiency trade-off.}
	\label{fig:pipline}
\end{figure*}

In this paper, we propose a simple yet effective filter pruning method, which can automatically obtain the pruning decision and the pruning ratio of each layer while considering the overall efficiency of the network, as shown in Fig. \ref{fig:pipline}.

To address the issue of the pruning indicator, we propose weight-dependent gates (W-Gates). Instead of designing a manual indicator or data-driven scale factors, we argue that the pruning decision should depend on the filter itself, in other words, it should be a learnable function of filter weights. Thus, we propose a type of novel weight-dependent gates to directly learn a mapping from filter weights to filter gates. W-Gates takes the weights of a convolutional layer as input to learn information from them, and outputs binary gates to open or close the corresponding filters automatically (0: close, 1: open) during the training process. Each W-Gates consists of a fully connected layer and binary activation, which is simple to implement and train given the pre-trained CNN model. More importantly, the output filter gates here are weights-dependent, which is the essential difference between the proposed W-Gates and conventional data-driven methods \cite{luo2020autopruner, huang2018data, liu2017learning}. 

To address the issue of the efficiency constraint, we propose a switchable Efficiency Module, which provides two options to cope with different scenarios, a Latency Prediction Network (LPNet) and FLOPs Estimation.
To prune the network under hardware constraint, we propose LPNet to predict the hardware latency of candidate pruned networks. LPNet takes the network encoding vector as input and outputs a predicted latency. 
After trained offline based on the latency data collected from the target hardware, the LPNet is able to provide latency guidance for network pruning.
Considering that hardware information is not always available, we add the FLOPs Estimation as an auxiliary unit in the Efficiency Module. For scenarios without hardware information, we can switch the Efficiency Module to the FLOPs Estimation and provide FLOPs guidance in a gradient manner for the pruning. 

With the proposed W-Gates and the switchable Efficient Module, we construct a differential pruning framework to perform filter pruning. We find that the output of W-Gates can be summed as a layer encoding to determine the pruning ratio of each layer, and all the layer encoding can constitute the network encoding vector of the candidate pruned architecture. Therefore, the network encoding is fed to the Efficiency Module to obtain the latency or FLOPs guidance. 
Then, Given the pre-trained CNN model and the LPNet, we can carry out network pruning in an end-to-end manner. To perform filter pruning during training, we define an efficiency-aware loss function that consists of the accuracy loss and the efficiency loss. The whole pruning framework is fully differentiable, allowing us to simultaneously impose the gradients of accuracy loss and efficiency loss to optimize the W-Gates of each layer. Retaining more filters will keep more information in the network, which is beneficial to the final accuracy. On the contrary, pruning more filters will improve the efficiency of the network inference. To this end, during training, the accuracy loss will pull the binary gates to more `1's (retaining more filters), while the efficiency loss pulls the binary gates to more `0's (pruning more filters). They compete against each other and finally obtain a compact network with the better accuracy-efficiency trade-off. 

It is worth mentioning that the pruning decision and the pruning ratio of each layer can all be obtained automatically, which involves little human partipication. As the input of the Efficiency Module is the network encoding, the pruning framework can carry out filter pruning with consideration of the overall network architecture, which is beneficial for finding optimal pruning ratios for different layers. The proposed W-Gates can be applied in various CNN-based computer vision tasks to compress and accelerate their backbones or the entire networks.

We evaluate our method on ResNets \cite{he2016deep}, MobileNet V2 \cite{Sandler_2018_CVPR}, and VGG \cite{simonyan2014very}. Comparing with uniform baselines, we consistently deliver much higher accuracy and lower latency. With lower latency, we achieve 1.09\%-1.33\% higher accuracy than ResNet34, 0.67\%-1.28\% higher accuracy than ResNet50, and 1.1\% higher accuracy than MobileNet V2 on ImageNet dataset. Compared to other state-of-the-art pruning methods \cite{Molchanov_2019_CVPR, he2019filter, lin2020hrank, guo2020dmcp, peng2019collaborative, ding2019centripetal, zhao2019variational, zhou2019accelerate, yu2019slimmable}, our method also produces superior results. 

The main contributions of our paper are three-fold:
\begin{itemize}
	\item We propose a kind of novel weight-dependent gates (W-Gates) to directly learn a mapping from filter weights to filter gates. W-Gates takes the convolutional weights as input and generates binary filter gates to prune or keep the filters automatically during training.
	\item A switchable Efficiency Module is constructed to predict the hardware latency or FLOPs of candidate pruned networks. Efficiency Module is fully differentiable with respect to W-Gates, allowing us to impose efficiency constraints based on gradients and obtain a compact network with a better accuracy-efficiency trade-off.
	\item The proposed method simultaneously addresses three challenges in filter pruning (pruning indicator, pruning ratio, efficiency constraint). Compared with state-of-the-art filter pruning methods, the proposed method achieves superior performance.
\end{itemize}

This paper extends the preliminary workshop paper \cite{li2020weight} in the following aspects. 1) We propose a switchable Efficiency Module to cope with various scenarios, so that the pruning framework can not only perform filter pruning under the latency constraint, but also under FLOPs constraints. 2) We generalize our idea of weight-dependent gates and latency predict network, enabling the application of our W-Gates from a single ResNet architecture to more types of CNN architectures, such as MobileNetv2, VGGNet, etc. 3) We conduct four important ablation studies to illustrate the rationality of the W-Gates design in our method. We show that W-Gates can indeed learn information from filter weights and make a good filter selection automatically, which helps the network training and obtains higher accuracy. 4) We investigate the effects of the key coefficient $\alpha$ in the proposed loss function on the final accuracy-latency trade-off. 5) We generalize our idea from a single dataset to more dataset scenarios. 6) We conduct visualization analysis on the pruned architectures. We show that W-Gates trends not to prune the layers with the downsample operations, and prunes more $3 \times 3$ convolutional layers than $1 \times 1$ layers. 

The rest of this paper is structured as follows. In Section II, we briefly review the related work. Section III details our proposed method. Experimental settings, ablation study, and results analysis are presented in Section IV followed by the conclusion in Section V. 

\section{Related Work}

A significant amount of effort has been devoted to deep model compression, such as matrix decomposition \cite{jia2017improving, yu2017compressing}, quantization \cite{hubara2016binarized, rastegari2016xnor},  knowledge distillation\cite{hinton2015distilling, zhang2021student}, compact architecture learning \cite{Sandler_2018_CVPR, zhang2018shufflenet}, neural network search \cite{guo2021jointpruning, guo2020single}, inference acceleration \cite{xu2018deepcache,li2021boosting}, and network pruning \cite{han2015deep, kang2019accelerator, li2019using}. Pruning is an intuitive and effective network compression method. Prior works devote to weights pruning. \cite{han2015deep} proposes to prune unimportant connections whose absolute weights are smaller than a given threshold, which achieves good performance on parameter compression. However, it is not implementation friendly and can not obtain faster inference without dedicated sparse matrix operation hardware. To tackle this problem, some filter pruning methods \cite{li2016pruning, kumar2020pruning, luo2018thinet} have been explored recently. These methods prune or sparse parts of the network structures (e.g., neurons, channels) instead of individual weights, so they usually require less specialized libraries to achieve inference speedup. In this paper, our work also falls into filter pruning. Next, we mainly discuss the works which are most related to our work.

\subsection{Manual-designed indicator}

Many excellent filter pruning methods based on manual-designed indicators have been proposed to compress large CNNs. \cite{li2016pruning} present an acceleration method for CNNs, where we prune filters from CNNs that are identified as having a small effect on the output accuracy.
\cite{he2017channel} proposes an iterative two-step algorithm to effectively prune each layer, by a LASSO regression based channel selection and least square reconstruction. \cite{luo2018thinet} formally establishes filter pruning as an optimization problem, and reveals that we need to prune filters based on statistics information computed from its next layer, not the current layer. The above two methods all prune filters based on feature maps. Different from the above data-driven methods, \cite{li2019exploiting} proposes a feature-agnostic method, which prunes filter based on a kernel sparsity and entropy indicator. \cite{he2018soft} proposes a soft filter pruning method, which prunes filters based on L2\_norm and updates the pruned model when training the model after pruning.
\cite{lin2018accelerating} introduces a binary global mask after each filter to dynamically and iteratively prune and tune the network, with the mechanism to recall filters that are mistakenly pruned in the previous iterations. \cite{he2019filter} proposes a Geometric Median based filter pruning method, which prunes filters with relatively less contribution and chooses the filters with the most replaceable contribution. 
These methods above need to manually set a global pruning ratio \cite{li2019exploiting, lin2018accelerating} or layer pruning ratios\cite{li2016pruning, he2017channel, luo2018thinet, he2019filter}, which is difficult to fully consider the redundancy of different layers. To solve this problem, \cite{wang2021convolutional} establishes a graph for each convolutional layer of a CNN and uses two quantities associated with the graph to obtain the redundancy and the pruning ratio of each layer. 
However, a common problem of these manual-designed pruning indicators is that they usually involve human participation, which makes them prone to be trapped in sub-optimal solutions.
To tackle this issue, our proposed method introduces Weight-dependent Gates to learn the pruning decision from filter weights automatically and cooperate with Efficiency Module to determine the pruning ratio of each layer, which involves little human participation.

\subsection{Data-driven Pruning Methods}

There are some other filter pruning methods that propose data-driven indicators. \cite{liu2017learning} imposes L1 regularization on the scaling factors in batch normalization (BN) layers to identify insignificant filters. Similarly, \cite{huang2018data} introduces scaling factors to scale the outputs of specific structures, such as neurons, groups, or residual blocks, and then do network pruning in an end-to-end manner. \cite{Luo_2020_CVPR} proposes to prune both channels inside and outside the residual connections via a KL-divergence based criterion to compensate for the weakness of limited data. \cite{guo2021gdp} proposes gates with differentiable polarization to control the on-and-off of each channel or whole layer block, which is essentially a kind of data-driven gates.
ManiDP\cite{tang2021manifold} explores a data-driven paradigm for dynamic pruning, which explores the manifold information of all samples in the training process and derives the corresponding sub-networks to preserve the relationship between different instances. ResRep\cite{ding2021resrep} proposes to re-parameterize a CNN into the remembering parts and forgetting parts, where the former learn to maintain the performance and the latter learn to prune.
\cite{ding2020prune} proposes a data-dependent soft pruning method, dubbed Squeeze-Excitation-Pruning (SEP), which does not physically prune any filters but selects different filters for different inputs. 
\cite{gong2022automatically} proposes a data-driven pruning indicator, in which the task-irrelevant channels are removed in a task-driven manner.
AutoPruner \cite{luo2020autopruner} proposed a data-driven channel selection layer, which takes the feature map of the previous layer as input and generates a binary index code for pruning. AutoPrunner \cite{luo2020autopruner} are most similar to our W-Gates. The major difference is that our W-Gates is weight-dependent, which directly learns information from filter weights instead of feature maps. The second difference is that AutoPrunner adopts a scaled sigmoid function as soft binarization to generate an approximate binary vector, while we adopt hard binarization (a variant of sign function) to obtain true binary gates during training.

\subsection{Quantization and Binary Activation.} 

Quantization\cite{rastegari2016xnor} proposes to quantize the real value weights into binary/ternary weights to yield a large amount of model size saving. \cite{soudry2014expectation} proved that good performance could be achieved even if all neurons and weights are binarized to $\pm$1. \cite{hubara2016binarized} proposes to use the derivative of the clip function to approximate the derivative of the sign function in the binarized networks. \cite{wu2019fbnet} relaxes the discrete mask variables to be continuous random variables computed by the Gumbel Softmax \cite{jang2016categorical} to make them differentiable.
\cite{liu2018bi} uses an identity shortcut to add the real-valued outputs to the next most adjacent real-valued outputs, and then adopt a tight approximation to the derivative of the non-differentiable sign function with respect to real-valued activation, which inspired our binary activation and gradient estimation in W-Gates.

\subsection{Resource-Constrained Compression}

Recently, real hardware performance has attracted more attention compared to FLOPs. AutoML methods \cite{he2018amc, yang2018netadapt} propose to prune filters iteratively in different layers of a CNN via reinforcement learning or an automatic feedback loop, which take real-time latency as a constraint. Some recent works \cite{wu2019fbnet, dai2019chamnet} introduce a look-up table to record the latency of each operation or each layer and sum them to obtain the latency of the whole network. This method is valid for many CPUs and DSPs, but may not for parallel computing devices such as GPUs. \cite{yang2019ecc} treats the hardware platform as a black box and creates an energy estimation model to predict the latency of specific hardware as an optimization constraint. \cite{guo2020dmcp} proposes DMCP to searching optimal sub-structure from unpruned networks under FLOPs constraint. 
\cite{gao2021network} trains a stand-alone neural network to predict sub-networks’ performance and then maximize the output of the network as a proxy of accuracy to guide pruning. \cite{liu2019metapruning} proposes PruningNet, which takes the network encoding vector as input and output weight parameters of the pruned network. These above works inspired our Efficiency Module and the LPNet training. We train the LPNet to predict the latency of the target hardware platform, which takes the network encoding vector as input and outputs the predicted latency.

\section{Proposed Method}

In this section, we introduce the proposed method which adopt an efficiency-aware approach to prune the CNN architecture under multiple constraints. 
Following the work \cite{liu2019metapruning}, we formulate the network pruning as a constrained optimization problem:

\begin{equation}
\begin{array}{*{20}{l}}
  {\arg \mathop {\min }\limits_{{w_c}} \ell \left( {c,{w_c}} \right)} \\ 
  {s.t.\; \; Inference\left ( c \right )\leq Const} 
\end{array}
\label{eq:optimization}
\end{equation}

where $\ell $ is the loss function specific to a given learning task, and $c$ is the network encoding vector, which is a set of the pruned network channel width $\left( {{c_1},{c_2}, \cdots ,{c_l}} \right)$. $Inference\left( c \right)$ denotes the real latency or FLOPs of the pruned network, which depends on the network channel width set $c$. ${w_c}$ means the weights of the remained channels and $Const$ is the given latency or FLOPs constraint.

To solve this problem, we propose an efficiency-aware network pruning framework, in which the weight-dependent gates is the key part. 
Firstly, we propose the Weight-dependent Gates Module (W-Gates), which is adopted to learn the information from filter weights and generate binary gates to determine which filters to prune automatically. The W-Gates works as adaptive pruning indicator. 
Then, a Efficiency Module is constructed to provide the latency or FLOPs constraint to the pruning indicator and guide the pruning ratios of each layer. As key parts of the efficient module, a Latency Prediction Network (LPNet) is trained offline to predict the real latency of a given architecture in specific hardware, and FLOPs Estimation is set to predict FLOPs for candidate architectures.
The two modules above complement each other to generate the pruning strategy and obtain the best CNN model under efficiency constraint.

\subsection{Weight-dependent Gates}

The convolutional layer has always been adopted as a black box, and we could only judge from the output what it has done. Conventional filter pruning methods mainly rely on hand-craft indicators or optimization-based indicators. They share one common motivation: they are essentially looking for a pruning function that can map filter weights to filter gates. However, their pruning function usually involves human participation.

\begin{figure}[t]
	\centering
	\includegraphics[width=7.5cm]{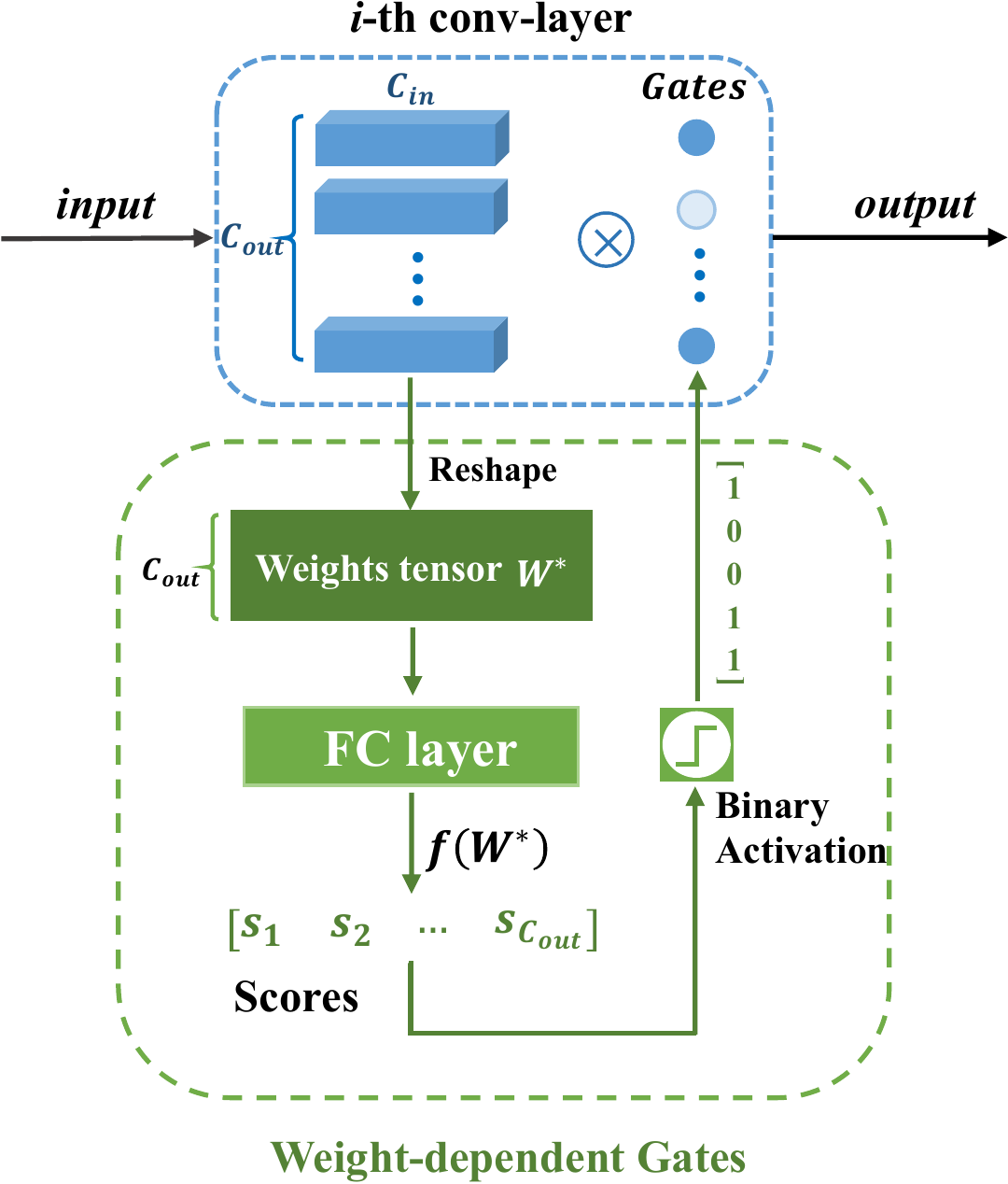}
	\centering
	\caption{The proposed Weight-dependent Gates. It introduces a fully connected layer to learn the information from the reshaped filter weights ${W^*}$ and generates a score for each filter. After binary activation, we obtain the filter gates (0 or 1) to open or close the corresponding filters automatically (0: close, 1: open). The gates are placed after the BN transform and ReLU. }
	\label{weights_learning}
\end{figure}

We argue that the gates should depend on the filters themselves, in other words, it is a learnable function of filter weights. Thus, instead of designing a manual indicator, we directly learn the pruning function from the filter weights, which is a direct reflection of the characteristics of filters.
To achieve the above goal, we propose the Weight-dependent Gates (W-Gates). W-Gates takes the weights of a convolutional layer as input to learn information, and output binary filter gates as the pruning function to remove filters adaptively.

\textbf{Filter Gates Learning.} 
Let $W \in {\mathbb{R}^{{C_l} \times {C_{l - 1}} \times {K_l} \times {K_l}}}$ denotes the weights of a convolutional layer, which can usually be modeled as $C_l$ filters and each filter ${W_i} \in {\mathbb{R}^{{C_{i - 1}} \times {K_l} \times {K_l}}},i = 1,2, \cdots ,{C_l}$. ${K_l}$ is the filter size of the $l$-th layer. To extract the information in each filter, a fully-connected layer, whose weights are denoted as $\overset{\lower0.5em\hbox{$\smash{\scriptscriptstyle\frown}$}}{W}  \in {\mathbb{R}^{({C_{l - 1}} \times {K_l} \times {K_l}) \times 1}}$, is introduced here.
Then, reshaped to two-dimensional tensor ${W^*} \in {\mathbb{R}^{{C_l} \times ({C_{l - 1}} \times {K_l} \times {K_l})}}$, the filter weights are input to the fully-connected layer to generate the score of each filter:
\begin{equation}
{s^r} = f\left( {{W^*}} \right)= {W^*}\overset{\lower0.5em\hbox{$\smash{\scriptscriptstyle\frown}$}}{W} ,
\end{equation}
where ${s^r} = \left[ {s_1^r,s_2^r, \ldots ,s_{{C_l}}^r} \right]$  denotes the score set of the filters in this convolutional layer.

To suppress the expression of filters with lower scores and obtain binary filter gates, we introduce the following activation function:
\begin{equation}
\sigma \left( x \right) = \frac{{{\mathop{\rm Sign}\nolimits} \left( x \right) + 1}}{2}.
\label{eq:binary}
\end{equation}

\begin{figure*}[t]  
	\centering
	\includegraphics[width=12.0cm]{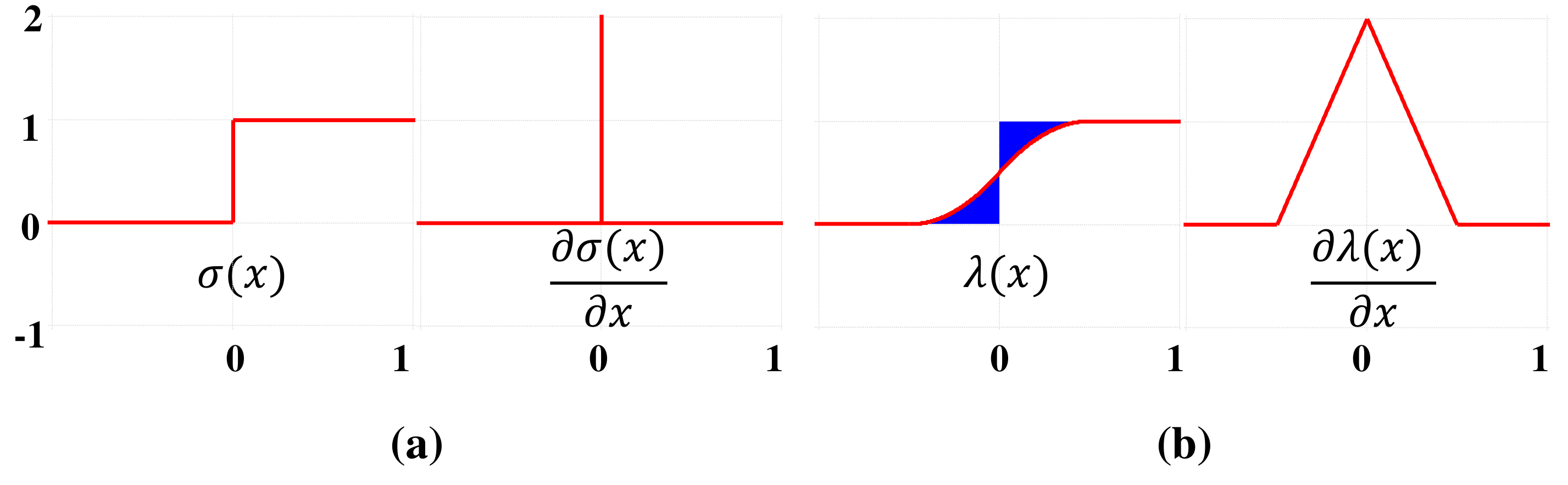}
	\centering
	\caption{(a) The proposed binary activation function and its derivative. (b) The designed differentiable piecewise polynomial function and its derivative, and this derivative is used to approximate the derivative of binary activation function in gradients computation. }
	\label{binary_function}
\end{figure*}

The curve of Eq. \eqref{eq:binary} is shown in Fig. \ref{binary_function}(a). We can see that after processing by the activation function, the negative scores will be converted to 0, and positive scores will be converted to 1. Then, we get the binary filter gates of this layer:
\begin{equation}
{{gates}^b} = \sigma \left( {{s^r}} \right) = \sigma \left( {f\left( {{W^*}} \right)} \right)
\label{eq:sign}.
\end{equation}

Different from the path binarization \cite{cai2019proxylessnas} in neural architecture search, in filter pruning tasks, the pruning decision should depend on the filter weights, in other words, our proposed binary filter gates are weights-dependent, as shown in Eq. \eqref{eq:sign}. Different from the time stepping controller \cite{yang2020dynamical} which is weight-dependent, the weight-dependent gates are proposed as pruning indicators.

Next, we sum the binary filter gates of each layer to obtain the layer encoding, and all the layer encodings form the network encoding vector $c$. The layer encoding here denotes the number of filters kept, which can also determine the pruning ratio of each layer.

\textbf{Gradient Estimation.} 
As can be seen from Fig. \ref{binary_function}, the derivative of function $\sigma \left ( \cdot  \right )$ is an impulse function, which cannot be used directly during the training process. Inspired by the recent Quantized Model works \cite{hubara2016binarized, liu2018bi}, specifically Bi-Real Net \cite{liu2018bi}, we introduce a differentiable approximation of the non-differentiable function $\sigma \left ( x  \right )$. The gradient estimation process is as follows:
\begin{equation}
\frac{{\partial L}}{{\partial {X_r}}} = \frac{{\partial L}}{{\partial {X_b}}}\frac{{\partial {X_b}}}{{\partial {X_r}}} =\frac{{\partial L}}{{\partial {X_b}}}\frac{{\partial \sigma \left( {{X_r}} \right)}}{{\partial {X_r}}} \approx \frac{{\partial L}}{{\partial {X_b}}}\frac{{\partial \lambda \left( {{X_r}} \right)}}{{\partial {X_r}}},
\end{equation}
where ${X_r}$ denotes the real value output ${s_i^r}$, ${X_b}$ means the binary output. ${\lambda \left( {{X_r}} \right)}$ is the approximation function we designed, which is a piecewise polynomial function:
\begin{equation}
\lambda \left( {{X_r}} \right) = \left\{ {\begin{array}{*{20}{l}}
  {0,\quad \quad \quad \quad \quad \quad if\;{X_r} <  - \frac{1}{2}} \\ 
  {2{X_r} + 2X_r^2 + \frac{1}{2},\;if\; - \frac{1}{2} \leq {X_r} < 0} \\ 
  {2{X_r} - 2X_r^2 + \frac{1}{2},\;\;if\;0 \leq {X_r} < \frac{1}{2}} \\ 
  {1,\quad \quad \quad \quad \quad \quad otherwise} 
\end{array}} \right.,
\end{equation}
and the gradient of above approximation function is:
\begin{equation}
\frac{{\partial \lambda \left( {{X_r}} \right)}}{{\partial {X_r}}} = \left\{ {\begin{array}{*{20}{l}}
  {2 + 4{X_r},\quad if\; - \frac{1}{2} \leq {X_r} < 0} \\ 
  {2 - 4{X_r},\quad if\;0 \leq {X_r} < \frac{1}{2}} \\ 
  {0,\quad \quad \quad \;otherwise} 
\end{array}} \right..
\label{gradient}
\end{equation}

As discussed above, we can adopt the binary activation function Eq. \eqref{eq:binary} to obtain the binary filter gates in the forward propagation, and then update the weights of the fully-connected layer with an approximate gradient Eq. \eqref{gradient} in the backward propagation.

\subsection{Efficiency Module}
\label{sec:LPNet}

With the proposed W-Gates, we can carry out filter scoring and selection based on the information in convolutional weights. However, to determine which structure is more efficient for inference, the hardware or FLOPs information is also necessary. For the motivation above, in this section, we introduce an Efficiency Module, which contains a Latency Prediction Network and a FLOPs Estimation unit to provide efficiency guidance for the proposed W-Gates. As shown in Fig. \ref{fig:pipline}, the Efficiency Module is a switchable module, in which the Latency Prediction Network is the main unit to select hardware-friendly architectures and the FLOPs Estimation is the substitute unit to provide FLOPs constraint.

\textbf{Latency Prediction Network.} Previous works on model compression aim primarily to reduce the parameters and calculations, but they does not always reflect the actual latency on hardware.
Therefore, some recent NAS-based methods \cite{liu2019metapruning, wu2019fbnet} pay more attention to adopt the hardware latency as a direct evaluation indicator than parameters and FLOPs. \cite{wu2019fbnet} proposes to build a latency lookup table to estimate the overall latency of a network, as they assume that the runtime of each operator is independent of other operators, which works well on many mobile serial devices, such as CPUs and DSPs. Previous works \cite{liu2019metapruning, wu2019fbnet} have demonstrated the effectiveness of this method. However, the latency generated by the lookup table is not differentiable with respect to the filter selection and the pruning ratio of each layer.

\begin{figure*}[t]
	\centering
	\includegraphics[width=12.0cm]{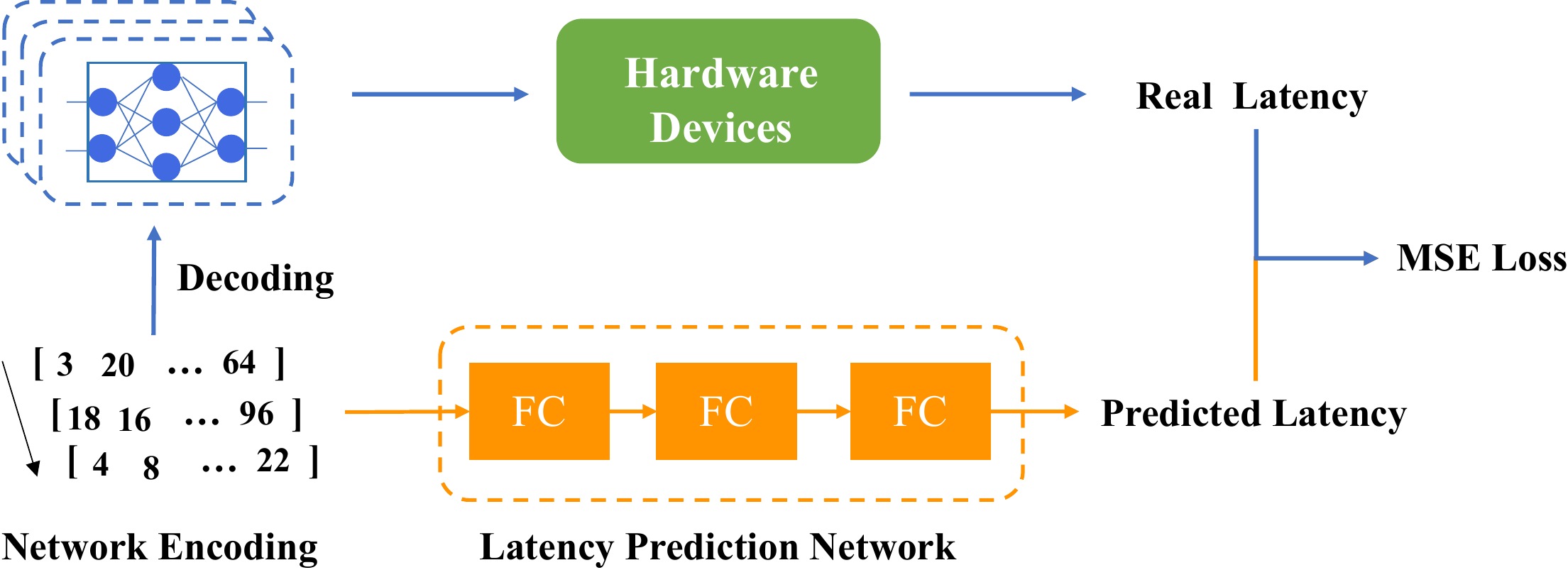}
	\centering
	\caption{The offline training process of LPNet. ReLU is placed after the first two FC layers. LPNet takes network encoding vectors as the input data and the measured hardware latency as the ground truth. Mean Squared Error (MSE) loss function is adopted here.}
	\label{latency}
\end{figure*}

To address the above problem, we construct LPNet to predict the real latency of the whole network or building blocks. The proposed LPNet is fully differentiable with respect to filter gates and the pruning ratio of each layer. 
As shown in Fig. \ref{latency}, the LPNet consists of three fully-connected layers, which takes a network encoding vector $c = \left( {{c_1},{c_2}, \ldots ,{c_L}} \right)$ as input and output the latency for specified hardware platform:
\begin{equation}
lat\left( c \right) = \operatorname{LPNet} \left( {{c_1},{c_2}, \ldots ,{c_L}} \right),
\end{equation}
where 
\begin{equation}
{c_l} = \sum_{i=1}^{C_{l}}gates_{i}^{b}
\end{equation}
in the pruning framework.

To pretrain the LPNet, we sample network encoding vectors $c$ from the search space and decode the corresponding network to test their real latency on specific hardware platforms. As shown in Fig. \ref{latency}, during the training process, the network encoding vector is adopted as input and the latency on specific hardware is used as the label. As there is no need to train the decoding network, it takes only a few milliseconds to get a latency label and the training of LPNet is also very efficient. For deeper building block architecture, such as ResNet50, the network encoding sampling space is huge. We choose to predict the latency of building blocks and then sum them up to get the predicted latency of the overall network, and this will greatly reduce the encoding sampling space. Besides, the LPNet of building blocks can also be reused across models of different depths and different tasks on the same type of hardware.

As a result, training such a LPNet makes the latency constraint differentiable with respect to the network encoding and binary filter gates shown in Fig. \ref{fig:pipline}. Thus we can use gradient-based optimization to adjust the filter pruning ratio of each convolutional layer and obtain the best pruning ratio automatically. 

\textbf{FLOPs Estimation.} Although the hardware latency is the most direct reflection of inference efficiency for the pruned models, it is not always available.
For scenarios with unknown hardware information, we add the FLOPs Estimation in the Efficient Module as an alternative unit to optimize FLOPs by gradient descent in the backward propagation. 

Let $c_{l-1}$ denotes the number of input channels of layer $l$, that is, the output channels of layer $l-1$. $c_{l}$ denotes the number of the output channels of layer $l$. Inspired by \cite{howard2017mobilenets, guo2020dmcp}, the FLOPs of convolutional layer $l$ in the pruned network can bu formulated as:
\begin{equation}
    F_{l}=M_{h}^{l}\cdot M_{w}^{l}\cdot K_{h}^{l}\cdot K_{w}^{l} \cdot c_{l-1} \cdot c_{l},
\end{equation}
in which, $M_{h}^{l}$ and $M_{w}^{l}$ are the height and the width of the input feature map in layer $l$, respectively. $K_{h}^{l}$ and $K_{w}^{l}$ denote the height and the width of the filter size, respectively. 

Among all layer types, convolutional layers are the primary performance hotspots \cite{xu2018deepcache}. Therefore, we focus on optimizing the FLOPs of convolutional layers when carrying out pruning. The FLOPs of a CNN with $L$ convolutional layers can be estimated as follows:
\begin{equation}
    Flops\left ( c \right ) =\sum_{l=1}^{L}F_{i}=\sum_{l=1}^{L}M_{h}^{l}\cdot M_{w}^{l}\cdot K_{h}^{l}\cdot K_{w}^{l} \cdot c_{l-1} \cdot c_{l},
\end{equation}
where $c_{0}=3$ (the number of channels of the input image) and ${c_l} = \sum_{i=1}^{C_{l}}gates_{i}^{b}$ in the pruning framework.

The FLOPs Estimation above is based on filter gates and training-free, which can be adopted to optimize the pruning process by gradient descent in the scenario with unknown hardware information.

\subsection{Efficiency-aware Filter Pruning}

If hardware information is available, the proposed method consists of three main stages. First, training the LPNet offline, as described in Section \ref{sec:LPNet}. With the pretrained LPNet, we can obtain the latency by inputting the encoding vector of a candidate pruned network. Second, pruning the network under latency constraint. We add W-Gates and the Efficiency Module (the LPNet) to a pretrained network to do filter pruning, in which the weights of LPNet are fixed. As shown in Fig. \ref{fig:pipline}, W-Gates learns the information from convolutional weights and generates binary filter gates to determine which filters to prune. Next, LPNet takes the network encoding of candidate pruned net as input and output a predicted latency to optimize the pruning ratio and filter gates of each layer. Then, the accuracy loss and the efficiency loss compete against each other during training and finally obtain a compact network with the best accuracy while meeting the latency constraint.
Third, fine-tuning the network. After getting the pruned network, a fine-tuning process with only a few epochs follows to regain accuracy and obtain a better performance, which is less time-consuming than training from scratch. 

Furthermore, to make a better accuracy-efficiency trade-off, we define the following efficiency-aware loss function:
\begin{equation}
\ell \left( {c,{w_c}} \right) = {\text{Acc}}\left( {c,{w_c}} \right) + \alpha \log \left( {1 + {\text{Eff}}\left( c \right)} \right),
\label{eq:loss}
\end{equation}
where ${\text{Acc}}\left( {c,{w_c}} \right)$ denotes the accuracy loss of an architecture with a network encoding $c$ and parameters ${{w_c}}$. ${{\text{Eff}}\left( c \right)}$ is the latency prediction $lat\left( c \right)$ or FLOPs estimation $Flops\left ( c \right )$ of the pruned architecture with network encoding vector $c$. The coefficient $\alpha$ can modulate the magnitude of the latency term. Such a loss function can carry out filter pruning tasks with consideration of the overall network structure, which is beneficial for finding optimal solutions for network pruning. In addition, this function is differentiable with respect to layer-wise filter choices $c$ and the number of filters, which allows us to use a gradient-based method to optimize them and obtain a better trade-off between accuracy and efficiency.

\section{Experiments}

In this section, we demonstrate the effectiveness of our method. First, we give a detailed description of our experiment settings. Next, we carry out four ablation studies on the ImageNet dataset to illustrate the effect of the key part W-Gates in our method. Then, we prune ResNet34, ResNet50, MobileNet V2, and VGG16 under latency constraint, and compare our method with several state-of-the-art filter pruning methods. Afterward, we prune VGG16 and ResNet56 under FLOPs constraint. Finally, we visualize the pruned architectures to explore what our method has learned from the network and what kind of architectures have a better trade-off between accuracy and efficiency.

\subsection{Experiment Settings}

We carry out experiments on the ImageNet ILSVRC 2012 dataset\cite{deng2009imagenet} and Cifar-10 dataset\cite{krizhevsky2009learning}.
ImageNet contains 1.28 million training images and 50,000 validation images, which are categorized into 1000 classes. The resolution of the input images is set to $224 \times 224$. Cifar-10 consisting of 50,000 training images, 10,000 test images, which are categorized into 10 classes. All images of Cifar-10 are cropped randomly into $32 \times 32$ with four paddings and horizontal flip is also applied.
All the experiments are implemented with PyTorch framework and networks are trained using stochastic gradient descent (SGD) with momentum set to 0.9. For ResNet34 and ResNet50, we adopt the same training scheme in \cite{he2016deep}.

To train the LPNet offline, we sample network encoding vectors $c$ from the search space and decode the corresponding network to test their real latency on one NVIDIA RTX 2080 Ti GPU as latency labels. 
For deeper building block architecture, such as ResNet50, the network encoding sampling space is huge. We choose to predict the latency of building blocks and then sum them up to get the predicted latency of the overall network, and this will greatly reduce the encoding sampling space. 
For the bottleneck building block of ResNet, we collect 170000 (c, latency) pairs to train the LPNet of bottleneck building block and 5000 (c, latency) pairs to train the LPNet of basic building block. We randomly choose 80\% of the dataset as training data and leave the rest as test data. We use Adam to train the LPNets and find the average test errors can quickly converge to lower than 2\%.

For the pruning and the fine-tuning processes, all networks are trained using stochastic gradient descent (SGD) with momentum set to 0.9. 
On ImageNet dataset, we respectively train the ResNets and MobileNet V2 for 120 epochs and 240 epochs as baselines. For ResNet34/50, we respectively set 40 and 80 epochs for the pruning and the fine-tuning processes. The initial learning rates of the two processes above are respectively set to $10^{-3}$ and $10^{-2}$. For MobileNet V2, we respectively set 120 and 80 epochs for the pruning and the fine-tuning processes. The initial learning rates of the two processes above are set to $10^{-3}$ and $10^{-1}$, respectively.
On Cifar-10 dataset, for all the models, we set 200 epochs for the pruning process and the fine-tuning process. The initial learning rates are set to $10^{-3}$ and $10^{-2}$, respectively.
On the two datasets, during the pruning and the fine-tuning processes, the leraning rates are all divided by 10 at 50\% and 75\% of the total number of epochs.

\subsection{Ablation Study on ImageNet}

The performance of our method is mainly attributed to the proposed Weight-dependent Gates (W-Gates). To validate the effectiveness of W-Gates, we choose the widely used architecture ResNet50 and conduct a series of ablation studies on ImageNet dataset. In the following subsections, we first explore the impact of our proposed W-Gates in the training process. Then, the impact of information learned from filter weights is studied. After that, we illustrate the impact of gate activation function by comparing binary activation with scaled sigmoid activation. Finally, we compare our W-Gates-based pruning method with several state-of-the-art gate-based methods \cite{liu2017learning, luo2018thinet, he2018soft}.

\begin{table}[t]
    \caption{This table compares the accuracy on ImageNet about ResNet50. ``W-Gates(warm-up)" denotes adding W-Gates to the ResNet50 after a warm-up period. ``W-Gates(pretrain)" means adding W-Gates to a pretrained ResNet50. ``W-Gates(constant input)" adopts a constant tensor input to replace filter weights in W-Gates. ``pretrain+same iters" means continuing to train the network with the same iterations as ``W-Gates(constant input)" and ``W-Gates(pretrain)".}
	\label{tab:ablation1}
	\begin{center}
		\begin{tabular}{ccccccccc}
			\hline
			~ \ \ \ & \ \ \ Top1-Acc \ \ \ & \ \ \ Top5-Acc \ \ \ \\
			\hline
			\hline
			\ \ \ \ ResNet50 baseline \ \ \ & \ \ \ 76.15\% \ \ \ & \ \ \ 93.11\% \ \ \ \\
			\hline
			\ pretrain + same iters & 76.58\% & 93.15\%  \\
			\hline
			\ W-Gates(constant input) & 75.32\% & 92.48\%  \\
			\hline
			\ \textbf{W-Gates(warm-up)} & \textbf{76.79\%} & \textbf{93.27\%} \\
			\hline
			\ \textbf{W-Gates(pretrain)} & \textbf{77.07\%} & \textbf{93.44\%} \\
			\hline
		\end{tabular}
	\end{center}
\end{table}

\subsubsection{Impact of W-Gates in the Training Process}
\label{sec:W-Gates effect}

In this section, to demonstrate the impact of W-Gates on filter selection, we add them to a pretrained ResNet50 to learn the weights of each convolutional layer and do filter selection during training. Then, we continue training and test the final accuracy. 

Two sets of experiments are set up to test the effect of W-Gates in different periods of the model training process. For one of them, we first train the network for 1/3 of total epochs as a warm-up period and then add W-Gates to the network to continue training. 
For the other experiment, we add W-Gates to a pretrained model and test if it could improve the training performance. As can be seen in Table \ref{tab:ablation1}, with the same number of iterations, adding W-Gates after a warm-up period can achieve 0.64\% higher Top-1 accuracy than baseline results. Moreover, equipping W-Gates to a pretrained network can continue to increase the accuracy by 0.92\%, which is 0.49\% higher than the result of adding the same number of iterations. These results show that adding the W-Gates to a well-trained network can make better use of its channel selection impact and obtain more efficient convolutional filters.

\subsubsection{Impact of Information from Filter Weights}

We are curious about such a question: \textit{Can the W-Gates really learn information from convolutional filter weights and give instruction to filter selection?} An ablation study is conducted to answer this question. First, we add the modules to a pretrained network and adopt constant tensors of the same size to replace the convolutional filter weights tensors as the input of W-Gates, and all the values of these constant tensors are set to ones.
Then we continue to train the network for the same number of iterations with ``W-Gates(pretrain)" in Table \ref{tab:ablation1}. 

From Table \ref{tab:ablation1}, we see that the W-Gates with a constant tensor as input achieves 1.75\% lower Top-1 accuracy than the W-Gates that input filter weights. The results above indicate that W-Gates can learn information from filter weights and do a good filter selection automatically, which contributes to the network training.

\subsubsection{Choice of Gate Activation Function}

\begin{table}[t]
	\caption{ This table compares two kinds of activation function: binary activation and scaled sigmoid activation, in which the scale factor is set to 4. (1G: 1e9)}
	\label{tab:ablation3}
	\begin{center}
		\begin{tabular}{cccc}
			\noalign{\smallskip}
			\hline
			~~ \ & \ Top1-Acc \  &  \ Top5-Acc \   &  \  FLOPs  \  \\ \hline \hline
			\ \ \ ResNet50 baseline  \ \ \ & 76.15\%  & 93.11\%  & 4.1G  \\ \hline
			W-Gates(sigmoid)              & 75.55\% & 92.62\% & 2.7G  \\ \hline
			\textbf{W-Gates(binary)}                                 & \textbf{76.01\%}  & \textbf{92.86\%}  & \textbf{2.7G}  \\ \hline
		\end{tabular}
	\end{center}
\end{table}
\setlength{\tabcolsep}{1.4pt}

There are two kinds of activation functions that can be used to obtain the filter gates, scaled sigmoid \cite{luo2020autopruner} and binary activation \cite{hubara2016binarized, liu2018bi}.
Previous methods adopt a sigmoid function or a scaled sigmoid function to generate an approximate binary vector. In this kind of method, a threshold needs to be set and the values smaller than it are set to 0. Quantized model works propose binary activation, which directly obtain a real binary vector and design a differentiable function to approximate the gradient of its activation function. We choose binary activation here as our gate activation function to generate the binary filter gates in W-Gates.

To test the impact of gate function choice in our method, we compare the pruning results of W-Gates with binary activation and W-Gates with scaled sigmoid. The scale factor of sigmoid $k$ is set to 4, which follows the setting in \cite{luo2020autopruner}. To keep only one factor changed, we do not adopt our LPNet to give the latency constraint, but simply imposes L1 regularization on the filter gates of each layer to obtain a sparse CNN model. As can be seen in Table \ref{tab:ablation3}, with the same FLOPs, W-Gates with binary activation achieves 0.46\% higher top-1 accuracy than W-Gates with k-sigmoid activation, which proves the binary activation is more suitable for our method.

\subsubsection{Compare with Other Gate-based Methods}

\begin{table}[t]
	\caption{ This table compares our W-Gates method with several state-of-the-art gate-based methods. For a fair comparison, we do not add LPNet to optimize the pruning process and only prune the network with L1-norm.}
	\label{tab:ablation4}
	\begin{center}
		\begin{tabular}{cccc}
			\noalign{\smallskip}
			\hline
			~~ \ \ \ & \ \ \ Top1-Acc \ \ \ &  \ \ \ Top5-Acc  \ \ \ &  \ \ \ FLOPs  \ \ \ \\ \hline \hline
			\ \ \ ResNet50 baseline  \ \ \ & 76.15\%  & 93.11\%  & 4.1G  \\ \hline
			SFP\cite{he2018soft}                                  & 74.61\%  & 92.87\% & 2.4G  \\ \hline
			Thinet-70\cite{luo2018thinet}                     & 75.31\%  & -        & 2.9G  \\ \hline
			Slimming\cite{liu2017learning}                    & 74.79\%  & 92.21\%  & 2.8G  \\ \hline
			\textbf{W-Gates}                                 & \textbf{76.01\%}  & \textbf{92.86\%}  & \textbf{2.7G}  \\ \hline
			\textbf{W-Gates}                                 & \textbf{75.74\%}  & \textbf{92.62\%}  & \textbf{2.3G}  \\ \hline
		\end{tabular}
	\end{center}
\end{table}
\setlength{\tabcolsep}{1.4pt}

To further examine whether the proposed W-Gates works well on filter pruning, we compare our method with state-of-the-art gate-based methods \cite{liu2017learning, luo2018thinet, he2018soft}. For Slimming \cite{liu2017learning}, there are no pruning results on ImageNet. To keep our setup as close to the original paper as possible, we adopt the original implementation publicly available and execute on Imagenet.
For a fair comparison, we do not add LPNet to optimize the pruning process but simply prune the network with L1-norm, which is consistent with other gate-based methods. The results for ImageNet dataset are summarized in Table \ref{tab:ablation4}. W-Gates achieves superior results than SFP\cite{he2018soft}, Thinet\cite{luo2018thinet} and Slimming\cite{liu2017learning}. The results show that the proposed Weight-dependent Gates can help the network to do a better filter self-selection during training.

\subsection{Pruning Results under Latency Constraint}

The inconsistency between hardware agnostic metrics and actual efficiency leads an increasing attention in directly optimizing the latency on the target devices. Taking the CNN as a black box, we train a LPNet to predict the real latency in the target device. For ResNet34 and ResNet50, we train two LPNets offline to predict the latency of basic building blocks and bottleneck building blocks, respectively. To fully consider all the factors and decode the sparse architecture, we add the factors of feature map size and downsampling to the network encoding vector. For these architectures with shortcut, we do not prune the output channels of the last layer in each building block to avoid mismatching with the shortcut channels.

\subsubsection{Pruning results on ResNet34}

\begin{table}[t]
	\caption{This table compares the Top-1 accuracy and latency on ImageNet about ResNet34. We set the same compression ratio for each layer as uniform baseline. The input batch size is set to 100 and the latency is measured using Pytorch on NVIDIA RTX 2080 Ti GPU. }
	\label{tab:ResNet34}
	\begin{center}
		\begin{tabular}{cccccc}
			\hline
			\multicolumn{3}{c}{\textbf{Uniform Baselines}} & \multicolumn{3}{c}{\textbf{W-Gates}} \\ \hline \hline
			\ FLOPs \ & \ Top1-Acc \  & \  Latency \  & \  FLOPs  \  & \ Top1-Acc \ & \  Latency \  \\ \hline
			3.7G (1X)    & 73.88\%    & 54.04ms    & -       & -         & -         \\ \hline
			2.9G         & 72.56\%    & 49.23ms    & 2.8G    & \textbf{73.76\%}   & \textbf{46.67ms}   \\ \hline
			2.4G         & 72.05\%    & 44.27ms    & 2.3G    & \textbf{73.35\%}   & \textbf{40.75ms}   \\ \hline
			2.1G         & 71.32\%    & 43.47ms    & 2.0G    & \textbf{72.65\%}   & \textbf{37.30ms}   \\ \hline
		\end{tabular}
	\end{center}
\end{table}

\begin{table}[t]
	\caption{This table compares the Top-1 accuracy and latency on ImageNet about ResNet50. The results show that, with the same FLOPs, our method outperforms the uniform baselines by a large margin in terms of accuracy and latency.}
	\label{tab:ResNet50}
	\begin{center}
		\begin{tabular}{cccccc}
			\noalign{\smallskip}
			\hline
			\multicolumn{3}{c}{\textbf{Uniform Baselines}} & \multicolumn{3}{c}{\textbf{W-Gates}} \\ \hline \hline
			\ FLOPs \ & \ Top1-Acc \  & \  Latency \  & \  FLOPs  \  & \ Top1-Acc \ & \  Latency \  \\ \hline
			4.1G (1X)    & 76.15\%    & 105.75ms    & -       & -         & -         \\ \hline
			3.1G         & 75.59\%    & 97.87ms    & 3.0G    & \textbf{76.26\%}   & \textbf{95.15ms}   \\ \hline			
			2.6G         & 74.77\%    & 91.53ms    & 2.6G    & \textbf{76.05\%}   & \textbf{88.00ms}   \\ \hline
			2.1G         & 74.42\%    & 85.20ms    & 2.1G    & \textbf{75.14\%}   & \textbf{80.17ms}   \\ \hline
		\end{tabular}
	\end{center}
\end{table}

We first employ the pruning experiments on a medium depth network ResNet34. ResNet34 consists of basic building blocks, each basic building block contains two $3 \times 3$ convolutional layers. We add the designed W-Gates to the first layer of each basic building block to learn the information and do filter selection automatically. LPNets are also added to the W-Gates framework to predict the latency of each building block. Then we get the latency of the whole network to guide the network pruning and optimize the pruning ratio of each layer. The pruning results are shown in Table \ref{tab:ResNet34}. We set the same compression ratio for each layer in ResNet34 as uniform baselines and prune the same layers with W-Gates. We measure the real hardware latency using Pytorch on NVIDIA RTX 2080 Ti GPU and the batch size of input images is set to 100. It can be observed that our method can save 25\% hardware latency with only 0.5\% accuracy loss on ImageNet dataset. With the same FLOPs, W-Gates achieves 1.1\% to 1.3\% higher Top-1 accuracy than uniform baseline, and the hardware latency is also lower, which shows that our W-Gates method can automatically prune and obtain the efficient architectures.

\subsubsection{Pruning results on ResNet50}

\begin{figure*}[t]  
	\centering
	\includegraphics[width=14.0cm]{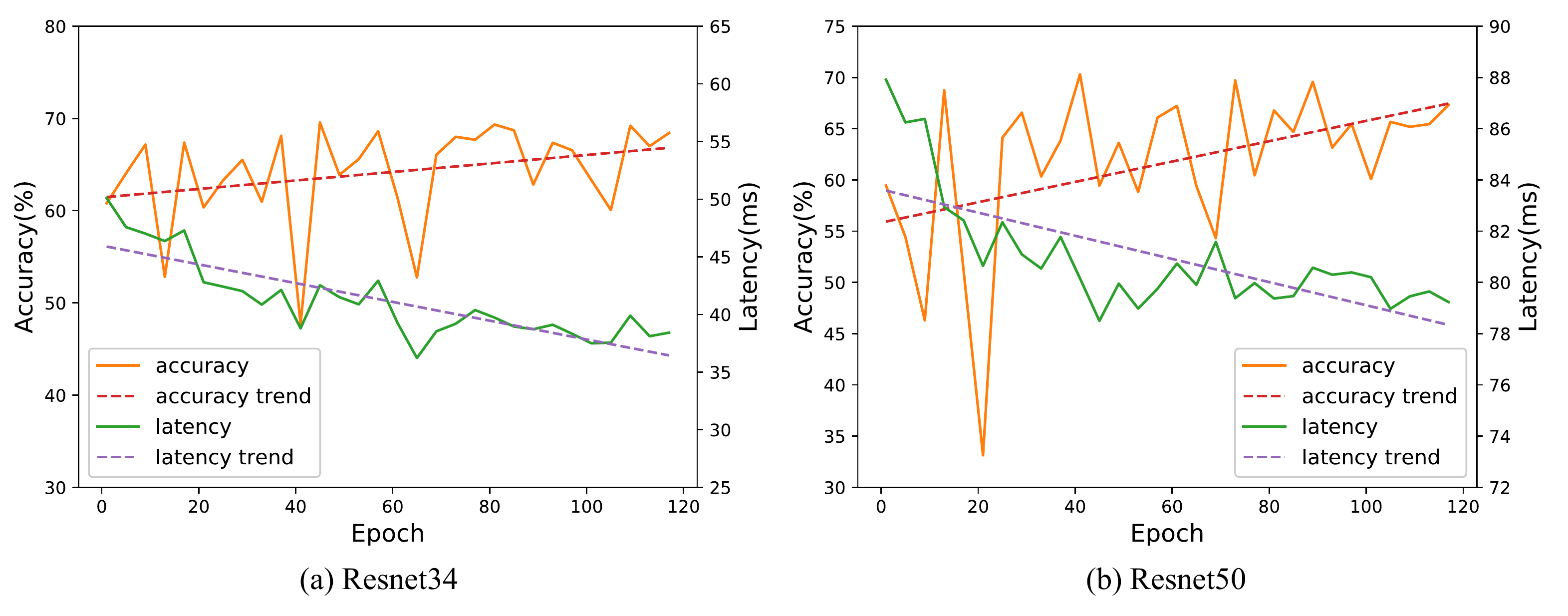}
	\centering
	\caption{ Two sets of pruning experiments on ResNet34 and ResNet50 on ImageNet. We set $\alpha$ in Eq.\eqref{eq:loss} to 1.5, which is used to modulate the magnitude of the latency term. In the training process, as the latency decreases, the accuracy first drops and then rises. These two figures show that W-Gates tries to find a better accuracy-latency trade-off via filter learning and selection in the training process. }
	\label{ResNet50_acc_lat}
\end{figure*}

For the deeper network ResNet50, we adopt the same setting with ResNet34. ResNet50 consists of bottleneck building blocks, each of which contains a $3 \times 3$ layer and two $1 \times 1$ layers. We employ W-Gates to prune the filters of the first two layers in each bottleneck module during training. 
The Top-1 accuracy and hardware latency of pruned models are shown in Table \ref{tab:ResNet50}. When pruning 37\% FLOPs, we can save 17\% hardware latency without notable accuracy loss. 
Then we plot two sets of pruning results on ResNet34 and ResNet50 in Fig. \ref{ResNet50_acc_lat}, in which $\alpha$ in Function \eqref{eq:loss} is set to 1.5 to modulate the magnitude of latency term. As can be seen from the accuracy and latency trends, during the training and selection process, as the latency decreases, the accuracy first drops and then rises. These two figures show that W-Gates tries to search for a better accuracy-efficiency trade-off via filter learning and selection in the training process.

\subsubsection{The Effect of Coefficient $\alpha$ on the Final Trade-off}

As the whole pruning framework is fully differentiable, we can simultaneously impose the gradients of accuracy loss and latency loss to optimize the W-Gates of each layer. The accuracy loss pulls the binary gates to more ones and the latency loss pulls the binary gates to more zeros. They compete against each other during training and finally obtain a compact network with a better accuracy-efficiency trade-off. 

In the proposed latency-aware loss function, the coefficient $\alpha$ can modulate the magnitude of the latency term. Now, we are interested in the effect of $\alpha$ for the final trade-off. Table \ref{tab:alpha} shows the $\alpha$ values and the corresponding latency of compact network when pruning ResNet34 on ImageNet. The input batch size is set to 100 and the latency is measured using Pytorch on NVIDIA RTX 2080 Ti GPU. It shows that the final actual latency will gradually decrease as alpha increases. For a given alpha, the latency will gradually decrease with the network training, and eventually converge to a best accuracy-latency trade-off. It can be seen that our method achieves 1.33X latency acceleration with only 0.53\% accuracy loss, which shows that W-Gates is very robust for the choice of $\alpha$ and can consistently deliver lower latency without notable accuracy loss.

\begin{table}[t]
	\caption{ The coefficient $\alpha$ values and the corresponding latency of compact network when pruning ResNet34 on ImageNet. The coefficient $\alpha$ can modulate the magnitude of latency term in the loss function. As shown in the results, the final actual latency will gradually decrease as alpha increases.}
	\label{tab:alpha}
	\begin{center}
		\begin{tabular}{ccccc}
			\hline
			\ \ $\alpha$ \ \ & \ \ 0 (Baseline) \ \  &  \ \  1.5  \ \  &  \ \  2.0  \ \  &  \ \ 3.0  \ \ \\ \hline
			\ \ Top1 Acc  \ \ & \ \ 73.88\% \ \ & \ \ 73.76\% \ \ & \ \ 73.35\% \ \ & \ \ 72.65\% \ \ \\ \hline
		    \ \ Latency  \ \ & \ \ 54.04ms \ \ & \ \ 46.67ms \ \ & \ \ 40.75ms \ \ & \ \ 37.30ms \ \ \\ \hline
		\end{tabular}
	\end{center}
\end{table}
\setlength{\tabcolsep}{1.4pt}

\begin{table}[!h]
	\begin{center}
	\caption{This table compares the Top-1 ImageNet accuracy of our W-Gates method and state-of-the-art pruning methods IENNP\cite{Molchanov_2019_CVPR}, FPGM\cite{he2019filter}, VCNNP\cite{zhao2019variational}, HRank\cite{lin2020hrank}, DMCP\cite{guo2020dmcp}, CCP\cite{peng2019collaborative}, RRBP\cite{zhou2019accelerate},  C-SGD\cite{ding2019centripetal}, S-MobileNet V2\cite{yu2019slimmable}, SRR-GR\cite{wang2021convolutional}, CLR-RNF\cite{lin2022pruning}, HAP\cite{yu2022hessian}, FilterSketch\cite{lin2021filter} on ResNet34, ResNet50, and MobileNet V2.}
	\label{tab:compare-with-SOTA}
		\begin{tabular}{c|lcc}
			\noalign{\smallskip}
			\hline
			\ \ Model \ \   & Methods \ \ \ & \ \ FLOPs \ \ & \ \ Top1 Acc \ \ \\ \hline \hline
			\multirow{4}{*}{\ \ \ ResNet34 \ \ \ } & IENNP\cite{Molchanov_2019_CVPR}  \ \ \  & \ \ \ 2.8G \ \ \ & \ \ \ 72.83\% \ \ \ \\ 
			& FPGM\cite{he2019filter}        & 2.2G  & 72.63\%  \\  \cline{2-4}
			&  \textbf{W-Gates}        & 2.8G  & \textbf{73.76\%}  \\  
			&  \textbf{W-Gates}        & 2.3G  & \textbf{73.35\%}  \\  \hline
			\multirow{21}{*}{ResNet50} 	
			& VCNNP\cite{zhao2019variational}        & 2.4G  & 75.20\%  \\ 
			& FPGM\cite{he2019filter}       & 2.4G  & 75.59\%  \\ 
			& FPGM\cite{he2019filter}        & 1.9G  & 74.83\%  \\  
			& IENNP\cite{Molchanov_2019_CVPR}         & 2.2G & 74.50\%    \\ 
			& DMCP\cite{guo2020dmcp} & 2.2G  & 76.20\%  \\
			& CCP\cite{peng2019collaborative} & 2.1G  & 75.50\%  \\
			& RRBP\cite{zhou2019accelerate}        & 1.9G & 73.00\%    \\ 
			& C-SGD-70\cite{ding2019centripetal}     & 2.6G  & 75.27\% \\ 
			& C-SGD-60\cite{ding2019centripetal}     & 2.2G  & 74.93\%  \\
			& C-SGD-50\cite{ding2019centripetal}     & 1.8G  & 74.54\%  \\ 
			& SRR-GR\cite{wang2021convolutional}        & 2.3G  & 75.76\% \\
			& CLR-RNF\cite{lin2022pruning}        & 2.5G  & 74.85\% \\
			& HAP\cite{yu2022hessian}   & 2.7G  & 75.12\%  \\
			& FilterSketch\cite{lin2021filter}        & 2.6G  & 75.22\% \\
			& FilterSketch\cite{lin2021filter}        & 2.2G  & 74.68\% \\
			& HRank\cite{lin2020hrank}               & 2.3G  & 74.98\%  \\ 
			& HRank\cite{lin2020hrank}               & 1.6G  & 71.98\%  \\ \cline{2-4}
			&  \textbf{W-Gates}   & 3.0G  &\textbf{76.26\%} \\ 
			&  \textbf{W-Gates}   & 2.4G  &\textbf{75.96\%}  \\ 
			&  \textbf{W-Gates}   & 2.1G  &\textbf{75.14\%}  \\ 
			&  \textbf{W-Gates}   & 1.9G  &\textbf{74.32\%}  \\ \hline
			\multirow{5}{*}{\ \ \ MobileNet V2 \ \ \ } 	& S-MobileNet V2\cite{yu2019slimmable}   \ \ \  & \ \ \ 0.30G \ \ \ & \ \ \ 70.5\% \ \ \ \\
			& S-MobileNet V2\cite{yu2019slimmable}     & 0.21G  & 68.9\%  \\ 
			& 0.75x MobileNetV2\cite{Sandler_2018_CVPR}    & 0.22G  & 69.8\%   \\  \cline{2-4}
			& \textbf{W-Gates}    & 0.29G  & \textbf{73.2\%}  \\  
			& \textbf{W-Gates}     & 0.22G  &\textbf{70.9\%}  \\  \hline
		\end{tabular}
	\end{center}
\end{table}
\setlength{\tabcolsep}{1.4pt}

\subsubsection{Comparisons with State-of-the-arts.}

In this section, we compare the proposed method with several state-of-the-art filter pruning methods, including manual-designed indicators (IENNP\cite{Molchanov_2019_CVPR}, FPGM\cite{he2019filter}, VCNNP\cite{zhao2019variational}, and HRank\cite{lin2020hrank}),  optimization-based methods (CCP\cite{peng2019collaborative}, RRBP\cite{zhou2019accelerate},  C-SGD\cite{ding2019centripetal}), and search-based methods (DMCP\cite{guo2020dmcp}, and S-MobileNet V2\cite{yu2019slimmable}), on ResNet34, ResNet50, and MobileNet V2. 
`W-Gates' in Table \ref{tab:compare-with-SOTA} denote our pruning results with different hardware latency and FLOPs. 
As there is no latency data provided in these works, we only compare the Top1 accuracy with the same FLOPs. 
It can be observed that, compared with state-of-the-art filter pruning methods, W-Gates achieves higher or comparable accuracy with the same FLOPs.  In particular, compared with C-SGD\cite{ding2019centripetal}, W-Gates achieves higher or comparable results (75.96\% vs  75.27\%, 75.14\%vs 74.93\% , 74.32\% vs 74.54\% ). Compared with Hrank\cite{lin2020hrank}, W-Gates achieve higher accuracy under 2.4G FLOPs (75.96\% vs 74.98\%). When the FLOPs of Hrank is compressed to 1.6G, its accuracy is only 71.98\%, which is 2.34\% lower than the 1.9G model of W-Gates, although their FLOPs gap is only 0.3G. 
This is because that W-Gates is trying to find a better trade-off between performance and efficiency during the pruning process. 

\subsubsection{Pruning Results on Cifar-10}

\begin{table}[t]
	\begin{center}
	\caption{This table compares the pruning results under latency constraint on Cifar-10 about VGG16 and ResNet56. Considering that the image size of Cifar-10 is very small (only $32\times 32$), we set the test batch size to $10^{4}$ to ensure the stability of the latency test results. (1M: 1e6)}
	\label{tab:cifar10}
		\begin{tabular}{c|lccc}
			\noalign{\smallskip}
			\hline
			\ \  Model \  \   & Methods \  \ & \ \ FLOPs \  \ & \  \ Top1 Acc \  \ & \ \ Latency \ \ \\ \hline \hline
			\multirow{4}{*}{VGG16} &  baseline   &  313M  &  93.72\%  &  391.1ms  \\ 
			& VCNNP\cite{zhao2019variational}        & 190M  & 93.18\%  &  - \\ 
			&  HRank\cite{lin2020hrank}        & 146M  & 93.43\%  & -  \\  
			&  \textbf{W-Gates}        &  \textbf{162M}  & \textbf{93.61\%}  & \textbf{245.0ms} \\  \hline
			\multirow{12}{*}{ResNet56} 	& 
			baseline        & 126.8M  & 93.85\%  &  501.8ms  \\ 
			& He et al.\cite{he2017channel}       &  63.4M  & 92.64\%  & -  \\ 
			& He et al.\cite{he2017channel}       &  62.0M  & 90.80\%  & -  \\
			& FPGM\cite{he2019filter}        & 60.1M  & 92.89\%  & - \\ 
			& FSDP\cite{gkalelis2020fractional}        & 64.4M  & 92.64\%  & - \\ 
			& PARI\cite{cai2021pruning}        & 60.1M  & 93.05\%  & - \\ 
            & CHIP\cite{sui2021chip}        & 34.8M  & 92.05\%  & - \\ 
			& ResRep\cite{ding2021resrep}        & 28.1M  & 92.66\%  & - \\ 
			& FilterSketch\cite{lin2021filter}        & 32.5M  & 91.20\%  & - \\ 
			& HRank\cite{lin2020hrank}        & 32.5M  & 90.72\%  & - \\  
			&  \textbf{W-Gates}   & \textbf{60.0M}  & \textbf{93.54\%}  &  \textbf{366.6ms} \\
			&  \textbf{W-Gates}   & \textbf{31.1M}  & \textbf{92.66\%}  &  \textbf{319.7ms}  \\ \hline
		\end{tabular}
	\end{center}
\end{table}
\setlength{\tabcolsep}{1.4pt}

In this section, we evaluate the proposed method on Cifar-10 dataset. The pruning experiments are conducted on two general models on Cifar-10, VGG16 and ResNet56. The results are shown in Table \ref{tab:cifar10}. It can be observed that the proposed W-Gates can also achieve good performance on Cifar-10 dataset, which outperforms state-of-the-art methods VCNNP\cite{zhao2019variational}, HRank\cite{lin2020hrank}, He et al.\cite{he2017channel}, FSDP\cite{gkalelis2020fractional}, FPGM\cite{he2019filter}, PARI\cite{cai2021pruning}, CHIP\cite{sui2021chip}, ResRep\cite{ding2021resrep}, and FilterSketch\cite{lin2021filter}. 
For the plain structure VGG16, W-Gates can reduce 37\% hardware latency and 48\% FLOPs with only 0.11\% accuracy loss. Similarly, for the ResNet56 architecture with shortcut connections, W-Gates can reduce 36\% hardware latency and 75\% FLOPs with only 1.2\% accuracy loss. 
Compared with state-of-the-art methods (\cite{he2017channel}, \cite{he2019filter}, \cite{lin2020hrank}, \cite{zhao2019variational}, \cite{ding2021resrep}, \cite{lin2021filter}, \cite{gkalelis2020fractional}, \cite{cai2021pruning}, \cite{sui2021chip}), W-Gates achieves higher or comparable results. 
Compared with ResRep\cite{ding2021resrep}, W-Gates has two advantages. First, in addition to FLOPs optimization, W-Gates can directly optimize the inference latency of CNNs on the target hardware platform in a gradient-based manner, which is not available in ResRep\cite{ding2021resrep}. Second, ResRep\cite{ding2021resrep} gradually increases the pruning ratio of each layer with a predefined filter pruning step (step=4), while W-gates can flexibly increase or decrease the pruning ratio of each layer based on the gradient in the backward propagation to obtain a more hardware-friendly CNN structure.
The results in Table \ref{tab:cifar10} can also prove the good generalization of the proposed W-Gates.

\subsection{Pruning Results under FLOPs Constraint}

For scenarios that the hardware information is unavailable, we can not directly optimize the pruning process under the latency constraint. With the proposed Efficiency Module, we can switch it to the FLOPs Estimation to cope well with these scenarios. In this section, we evaluate the proposed pruning framework under FLOPs constraint. The experiments are conducted on CIFAR10. 

Table \ref{tab:flops} summarizes the achieved improvement via applying FLOPs constraint on filter pruning. Our primary observation is that our method can have substantial FLOPs reduction with negligible accuracy loss compared to other state-of-the-art methods. 
For the plain architecture VGG16, our method reduces 43.8\% FLOPs with only 0.23\% accuracy loss. For a deeper model ResNet56, our method achieves 78.6\% FLOPs reduction with only 2.25\% accuracy loss.  
From the results we can also see that the accuracy in Table \ref{tab:flops} is slightly lower than the performance of pruning under latency constraint in Table \ref{tab:cifar10}. The reason is that, FLOPs optimization may mainly focuses on pruning the shallow layers of the network (the primary FLOPs hotspots), which may damage the diversity of low-level features extracted by the shallow layers of the network. In contrast, the latency optimization pays more attention to the hardware friendliness of the whole pruned architecture, rather than only focusing on local parts of the network. Fortunately, compared with state-of-the-art methods \cite{zhao2019variational, lin2020hrank, he2017channel}, our method under FLOPs constraint can also achieve superior or comparable performance, suggesting that it is a good auxiliary constraint. 

\begin{table}[t]
	\begin{center}
	\caption{This table compares the pruning results under FLOPs constraint on Cifar-10 about VGG16 and ResNet56. }
	\label{tab:flops}
		\begin{tabular}{c|lcc}
			\noalign{\smallskip}
			\hline
			\ \  Model \  \   & Methods \  \ & \ \ FLOPs \  \ & \  \ Top1 Acc \  \  \\ \hline \hline
			\multirow{4}{*}{VGG16} &  baseline   &  313M  &  93.72\%   \\ 
			& VCNNP\cite{zhao2019variational}        & 190M  & 93.18\%  \\ 
			&  HRank\cite{lin2020hrank}        & 146M  & 93.43\%  \\  
			&  \textbf{W-Gates}        &  \textbf{176M}  & \textbf{93.49\%}  \\  \hline
			\multirow{4}{*}{ResNet56} 	& 
			baseline        & 126.8M  & 93.85\%   \\ 
			& He et al.\cite{he2017channel}       &  62.0M  & 90.80\%  \\ 
			& HRank\cite{lin2020hrank}        & 32.5M  & 90.72\%   \\  
			&  \textbf{W-Gates}   & \textbf{27.1M}  & \textbf{91.60\%}  \\ \hline
		\end{tabular}
	\end{center}
\end{table}
\setlength{\tabcolsep}{1.4pt}

\begin{table}[t]
	\begin{center}
	\caption{This table compares the top-1 accuracy of the pruned models before and after the fine-tuning process under FLOPs constraint. }
	\label{tab:flops_acc}
		\begin{tabular}{c|ccc}
			\noalign{\smallskip}
			\hline
			\ \  Model \  \   & Top-1 Acc (before) \  \ & \ \ Top-1 Acc (after) \  \ & \  \ $\Delta $ \  \  \\ \hline 
			VGG16 &   92.82\%      &  93.49\%  &  0.67\% \\ 
			ResNet56 &   91.20\%     &  91.60\%  &  0.40\%  \\  \hline
		\end{tabular}
	\end{center}
\end{table}
\setlength{\tabcolsep}{1.4pt}

Table \ref{tab:flops_acc} shows the top-1 accuracy of the pruned models before and after the fine-tuning process under FLOPs constraint. It is observed that when the pruning process is just completed, W-Gates has achieved a good performance on VGG16 and ResNet56. After the fine-tuning process, the Top-1 accuracy of VGG16 and ResNet56 can be further regained by 0.67\% and 0.40\%, respectively. These results indicate that the performance of W-Gates mainly comes from the pruning stage.
The reason is that, with the proposed efficiency-aware loss function, W-Gates can carry out filter pruning tasks with consideration of the overall network structure, which is beneficial for obtain a better trade-off between accuracy and efficiency and finding optimal solutions for network pruning.

\subsection{Visualization Analysis}

In the filter pruning process, we are curious about what W-Gates have learned from the network and what kind of architectures have a better trade-off in accuracy-latency. In visualizing the pruned architectures of ResNet34 and ResNet50, we find that the W-Gates did learn something interesting.

\begin{figure}[t]  
	\centering
	\includegraphics[width=8.0cm]{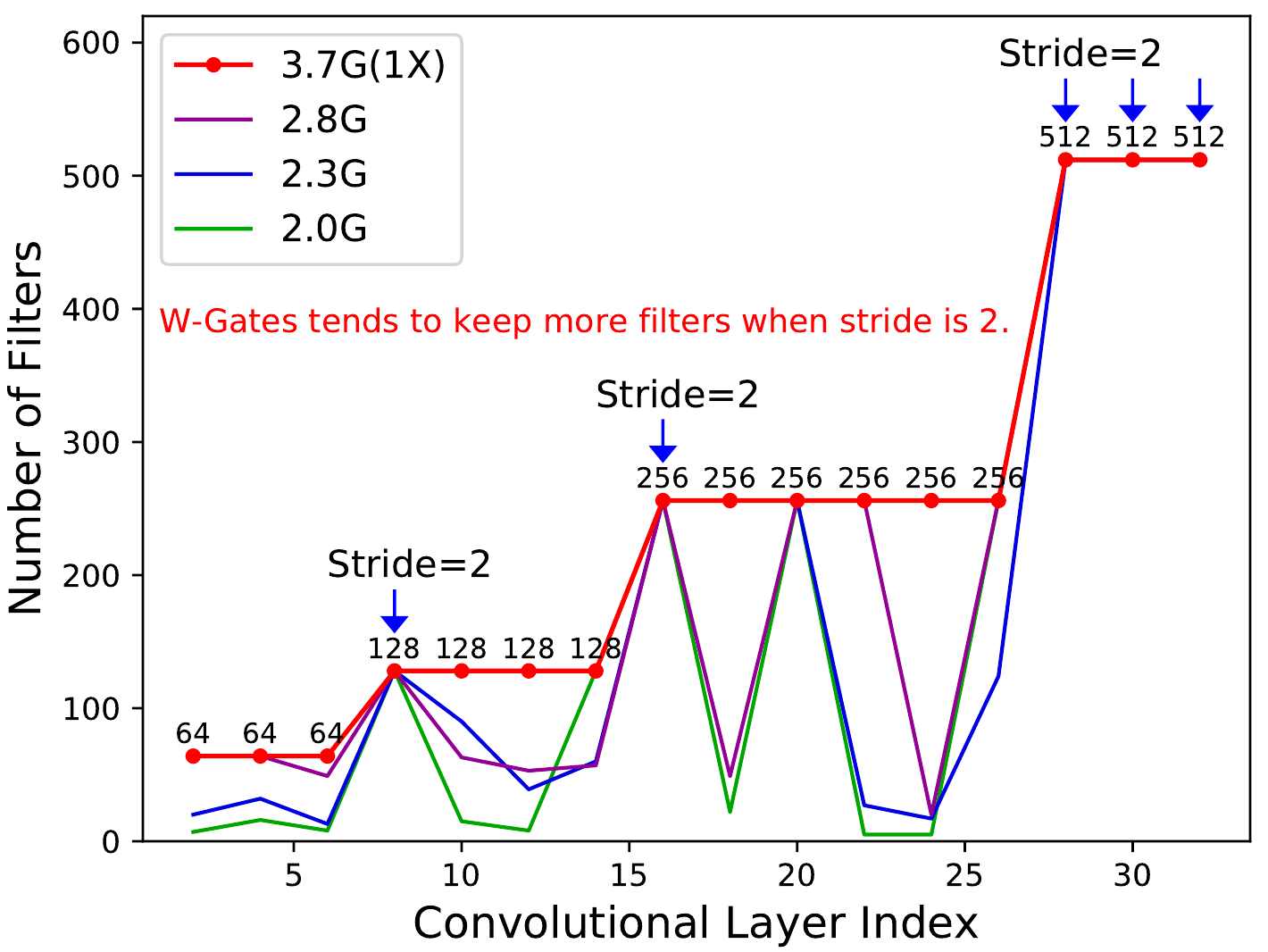}
	\centering
	\caption{Visualization of the pruning results on ResNet34. For such an architecture consisting of basic building blocks, our W-Gates method learns to keep more filters where there is a downsampling operation.}
	\label{ResNet34_visual}
\end{figure}

\begin{figure}[t]
	\centering
	\includegraphics[width=8.0cm]{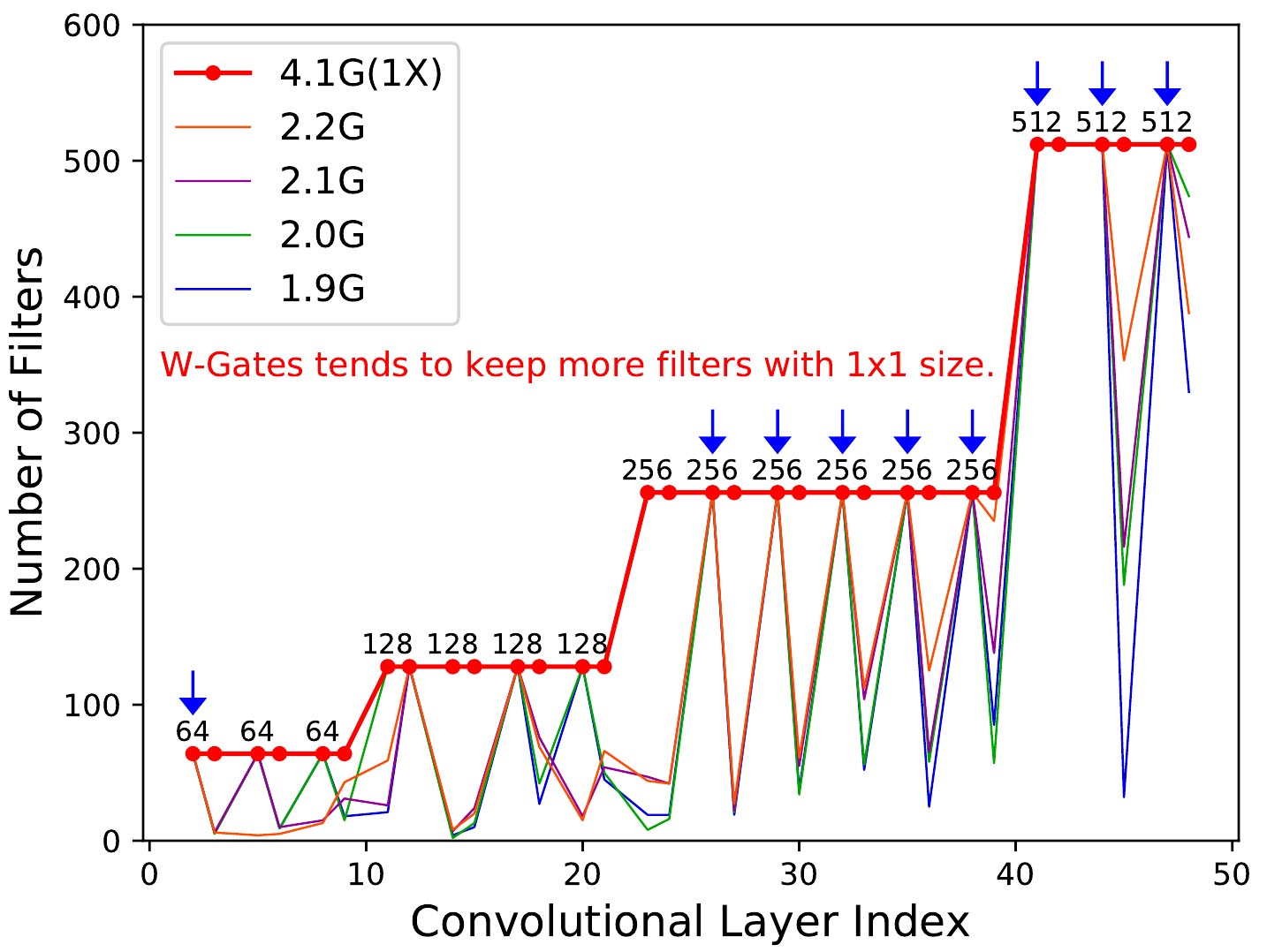}
	\centering
	\caption{Visualization of the pruning results on ResNet50. We prune filters of the first two layers in each bottleneck building block. For such an architecture consisting of bottleneck building blocks, our W-Gates method learns that the $3 \times 3$ convolutional layers have larger information redundancy than $1 \times 1$ convolutional layers, and contribute more to the hardware latency. }
	\label{ResNet50_visual}
\end{figure}

The visualization of ResNet34 pruned results under latency constraint on ImageNet is shown in Fig. \ref{ResNet34_visual}. It can be observed that for the architecture ResNet34 consisting of basic building blocks, W-Gates trends not to prune the layer with the downsampling operation, although a large ratio of filters in the other layers have been pruned. This is similar to the results in \cite{liu2019metapruning} on MobileNet, but is more extreme in our experiments on ResNet34.  It is possibly due to that the network needs more filters to retain the information to compensate for the loss of information caused by feature map downsampling.

However, for ResNet50 architecture consisting of bottleneck building blocks, the phenomenon is different. As can be seen in Fig. \ref{ResNet50_visual}, the W-Gates trends not to prune the $1 \times 1$ convolutional layers and prune the $3 \times 3$ convolutional layers with a large ratio. It shows that W-Gates learn automatically that the $3 \times 3$ convolutional layers have larger information redundancy than $1 \times 1$ convolutional layers, and contribute more to the hardware latency.


\section{Conclusions}

In this paper, we propose a novel filter pruning method to address the problems on pruning indicator, pruning ratio, and platform constraint at the same time. We first propose weight-dependent gates to learn the information from convolutional weights and generate novel weights-dependent filter gates. Then, we construct a switchable Efficiency Module to predict the latency or FLOPs of the candidate pruned networks and provide efficiency constraint for the weight-dependent gates during the pruning process.
The entire framework is fully differentiable with respect to filter choices and pruning ratios, which can be optimized by a gradient-based method to achieve better pruning results.

However, the proposed method also has some limitations. On the one hand, for the ResNet architecture with shortcut connections, we did not prune the output channels of the last layer in each building block, which left some compression space unused. In our future study, we will try to find a better way to solve this issue. On the other hand, we only focused on classification tasks in this paper and did not apply the proposed method to other types of applications. In future work, we will try to transfer the proposed method to other application scenarios, such as object detection, semantic segmentation, etc.


\bibliography{egbib}
\bibliographystyle{IEEEtran}

\end{document}